\pdfoutput=1

\documentclass[11pt]{article}

\usepackage{acl}

\usepackage{times}
\usepackage{latexsym}
\usepackage{graphicx}
\usepackage{amsmath}
\usepackage{hyperref}
\hypersetup{hidelinks,
	colorlinks=true,
	allcolors=black,
	pdfstartview=Fit,
	breaklinks=true}
\usepackage{multirow}
\usepackage{color}
\usepackage{enumitem}
\setlist[itemize]{leftmargin=*}

\usepackage[T1]{fontenc}

\usepackage[utf8]{inputenc}

\usepackage{microtype}

\title{SelF-Eval: Self-supervised Fine-grained Dialogue Evaluation}

\author{Longxuan Ma \and  Ziyu Zhuang \and Weinan Zhang\thanks{*Corresponding author} \and Mingda Li \and Ting Liu\\
Research Center for Social Computing and Information Retrieval, Harbin Institute of Technology\\
\texttt{lxma,zyzhuang,wnzhang,mdli,tliu@ir.hit.edu.cn} 
}

\begin{document}
\maketitle
\begin{abstract}
This paper introduces a novel \textbf{Sel}f-supervised \textbf{F}ine-grained Dialogue \textbf{Eval}uation framework (\textbf{SelF-Eval}). The core idea is to model the correlation between turn quality and the entire dialogue quality. We first propose a novel automatic data construction method that can automatically assign fine-grained scores for arbitrarily dialogue data. Then we train \textbf{SelF-Eval} with a multi-level contrastive learning schema which helps to distinguish different score levels. Experimental results on multiple benchmarks show that SelF-Eval is highly consistent with human evaluations and better than the state-of-the-art models. We give a detailed analysis of the experiments in this paper. Our code is available on GitHub.
\end{abstract}%

\section{Introduction}

Dialogue systems (DS) aim to satisfy human needs \cite{DBLP:journals/jzusc/ShumHL18,DBLP:conf/ijcai/Yan18,DBLP:journals/ftir/GaoGL19} such as information, communication, entertainment, etc. Appraising the quality of the DS responses reflects the system’s capability and provides insights into required further improvements \cite{DBLP:conf/sigdial/FinchC20,DBLP:journals/corr/abs-1905-04071}. Among the commonly used evaluation metrics, human evaluation is of high reliability but expensive to conduct, automatic metrics used in language generation (Perplexity \cite{DBLP:conf/nips/BengioDV00}) or machine translation (BLEU \cite{DBLP:conf/acl/PapineniRWZ02}, ROUGE \cite{lin2004rouge}, etc.) are easy to conduct but ineffective to reflect the dialogue quality \cite{DBLP:conf/emnlp/LiuLSNCP16,DBLP:conf/emnlp/NovikovaDCR17}. Therefore, researchers have made great efforts to find more reliable automatic evaluation metrics that are highly correlated with human evaluation \cite{DBLP:conf/acl/LoweNSABP17,DBLP:conf/aaai/TaoMZY18,DBLP:conf/sigdial/MehriE20}.

\begin{figure}[t]
\centering
\includegraphics[width=3in]{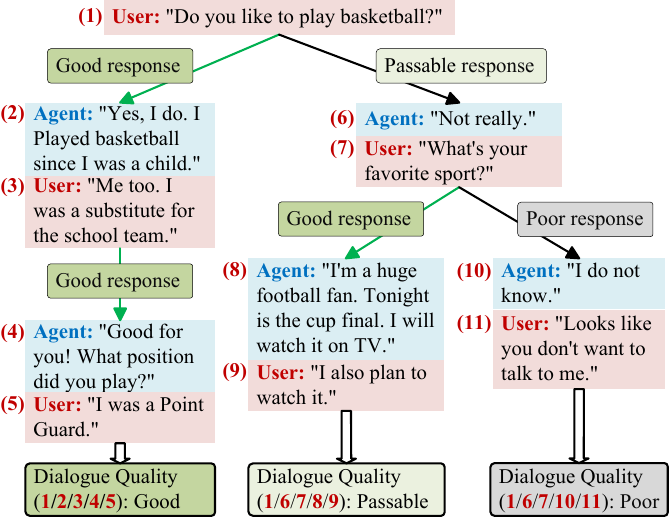}
\caption{The fine-grained relationships between turn-level and dialogue-level quality.}
\vspace{-0.5cm}
\label{Example}
\end{figure}

The current automatic dialogue evaluation metrics leverage semantic information \cite{DBLP:conf/emnlp/HuangYQLL20,DBLP:conf/acl/MehriE20,DBLP:conf/acl/YeLHLL20} to measure dialogue quality. For example, when evaluating the response quality, they either compute the semantic similarity between dialogue context and the generated response \cite{DBLP:conf/emnlp/XuDKR18,DBLP:conf/aaai/TaoMZY18,DBLP:conf/naacl/DziriKMZ19,DBLP:journals/corr/abs-1904-10635} or measure the soft semantic overlap between ground-truth response and the model-generated one \cite{DBLP:conf/acl/LoweNSABP17,DBLP:conf/naacl/XuJLRWWWW18,DBLP:conf/emnlp/ZhaoPLGME19,DBLP:conf/iclr/ZhangKWWA20,DBLP:conf/nips/YuanNL21}. When evaluating \cite{DBLP:conf/acl/ZhangCDZFL020} the overall dialogue quality, they either learn a dialogue-level representation for rating directly \cite{DBLP:conf/acl/MesgarBG20,DBLP:conf/acl/ZhangCDZFL020} or calculate the score with the help of other indirect assists \cite{DBLP:conf/sigdial/MehriE20}. However, recent studies \cite{DBLP:conf/sigdial/MehriE20,DBLP:conf/emnlp/SaiDSMK21,DBLP:journals/corr/abs-2106-03706} show that current models can only work well for measuring the response or evaluating the entire dialogue. They could not perform well in both situations at the same time. It means that the dialogue representation they learned \cite{DBLP:conf/acl/ZhangCDZFL020} could not reflect both turn quality and the entire dialogue quality. 

The dialogue quality is affected by all turns' qualities in it \cite{DBLP:conf/interspeech/GopalakrishnanH19} and this effect is accumulated in a multi-turn dialogue \cite{DBLP:journals/corr/abs-2103-01287}. Figure \ref{Example} shows these quality correlations between turns and dialogue. Each turn is marked with a serial number. Three dialogue examples are starting with the same user turn (1). The left example (1/2/3/4/5) shows that two good agent responses result in good overall quality. The middle (1/6/7/8/9) and right (1/6/7/10/11) examples show how lower-quality agent responses result in different dialogue qualities (passable or poor). The current open-domain dialogue evaluation methods fail to model the fine-grained correlations between turn quality and dialogue quality, which entail a poor dialogue representation for evaluation. 

In this paper, we introduce an evaluation method that explicitly models the correlations between turn quality and dialogue quality. Specifically, we aim to learn a dialogue representation that can reflect each turn's contribution, so that the evaluation score obtained by this representation aligns turn quality with the dialogue quality. To this end, we need to first obtain large amounts of dialogue data that reflects fine-grained correlations between turns quality and dialogue quality, then train an evaluation model to measure the fine-grained correlations. The contributions of this paper are:

\vspace{-0.1cm}
\begin{itemize}
\item To the best of our knowledge, we are the first to explicitly model the fine-grained correlation between turns and the entire dialogue for open-domain dialogue evaluation.
\vspace{-0.2cm}
\item We introduce a simple but effective data construction method to align the turn-level quality with the overall dialogue quality. We design a \textbf{Sel}f-supervised \textbf{F}ine-grained Dialogue \textbf{Eval}uation model (SelF-Eval) with a multi-level contrastive learning (MLCL) method. Our code and data are publicly available: \href{https://github.com/royny/SelF-Eval}{https://github.com/royny/SelF-Eval.}
\vspace{-0.2cm}
\item Experiments on multiple benchmarks show that SelF-Eval: 1) can evenly distinguish different replacement levels; 2) builds the correlations between turn qualities and dialogue qualities; 3) gets better correlation scores with human ratings than the state-of-the-art (SOTA) models. 
\end{itemize}

\section{Related Work}
We first survey evaluation metrics in open-domain dialogue (sections 5.1 and 5.2), then compare related work in task-oriented dialogue (section 5.3).

\subsection{Calculation of Semantic Overlap}
In this category, metrics are designed to measure the semantic similarity between the generated response and the dialogue context \cite{DBLP:conf/emnlp/XuDKR18,DBLP:conf/aaai/TaoMZY18,DBLP:journals/corr/abs-1904-10635,DBLP:conf/acl/PangNHZLT20} or soft semantic overlap between the generated response and the reference response \cite{DBLP:conf/acl/LoweNSABP17,DBLP:conf/naacl/XuJLRWWWW18,DBLP:conf/iclr/ZhangKWWA20,DBLP:conf/emnlp/ZhaoPLGME19,DBLP:conf/nips/YuanNL21}. \citet{DBLP:conf/naacl/DziriKMZ19} presented interpretable metrics for evaluating topic coherence by making use of distributed sentence representations. COMET \cite{DBLP:conf/emnlp/ReiSFL20} evaluated machine translation quality with a pre-trained model by minimizing the distance of the hypothesis with both reference and source text. The most similar work to ours is from \citet{DBLP:conf/acl/YeLHLL20} that measures quantifiable coherence scores. \textit{The differences between their work and ours are: 1) they focus on turn-level evaluations while we aim to evaluate both turn and dialogue-levels. Their method models the relationship between dialogue context and response while we model the fine-grained correlations between turns and the entire dialogue; 2) their method relies on the multi-level human annotations for dialogue quality while our method is free from these constrain. SelF-Eval is trained in a self-supervised manner, using synthetic dialogue data and automatically annotated scores.} 

\subsection{Regression to a Reference Score}
In this category, metrics learn to evaluate dialogue with scores that represent pre-defined dialogue attributes. BLEURT \cite{DBLP:conf/acl/SellamDP20} trained a BERT model with synthetic data and fine-tuned it on human ratings. GRADE \cite{DBLP:conf/emnlp/HuangYQLL20} introduced dialogue topic transitions for coherence evaluation. USR \cite{DBLP:conf/acl/MehriE20} leveraged RoBERTa and a regression model to approximate the specific scores rated by annotators. \citet{DBLP:conf/sigdial/MehriE20} computed the log-likelihood of DialoGPT generating predefined positive or negative comments as the score. The most similar work to ours in this category is from \citet{DBLP:conf/acl/MesgarBG20}, they utilized dialogue act labels to help dialogue level representation learning and assist the performance of dialogue-level coherence evaluation. \textit{The difference between their work and ours are: 1) they use dialogue act to assist the dialogue representation learning in a multi-task learning framework while we only use dialogue information; 2) their method only measures the coherence of dialogue while ours measures multiple attributes of dialogue; 3) their method aims to distinguish good samples from bad ones while ours can assign fine-grained scores for each sample.}

\subsection{Related Work in Task-oriented DS}
Besides the open-domain dialogue, there is also work in task-oriented dialogue similar to ours. They assume that users start a dialogue with a task-sensitive patience budget and the dialogue is finished by users when the task is completed or the budget runs out. Their model estimates user satisfaction at each turn and consumes some remaining budget. \textit{The differences between their work and ours are: 1) they focus on task-oriented dialogue with explicitly dialogue purpose while ours are open-domain dialogue with no such goals; 2) they need expensive training and data collecting pipeline while ours do not need; 3) they set up an overall budget which is consumed during the dialogue while we learn a dialogue-level representation for evaluation.} When adopting our method to evaluate task-oriented dialogues, information such as intents and request types is required to determine task completion. Our model will require further improvements to utilize this information.

\section{Our Proposed Method}

\subsection{Problem Statement}

Given an $n$ rounds (2*$n$ turns) dialogue \textbf{D} = [$A_1$,$B_1$,$A_2$,$B_2$,...,$A_{n}$,$B_{n}$] where $A_i$/$B_i$ represents the $i$-th ($i\in\{1,2,...,n\}$) turn from human-A/machine-B, respectively. The dialogue evaluation model takes \textbf{D} as input and outputs a quality score (a scalar value) for it. 

\subsection{Data Construction}
We need training data with quantitative annotation on both turn-level and dialogue-level (Figure \ref{Example}). However, only very few dialogue data today have these kinds of labels \cite{DBLP:conf/interspeech/GopalakrishnanH19} and the models trained with this kind of data are restricted by domain adaptability and generality. Inspired by previous work \cite{DBLP:conf/acl/MesgarBG20,DBLP:conf/acl/ZhangCDZFL020}, we adopt a replacement strategy that perturbs a dialogue at the semantic level. In this strategy, the easily accessible human-human dialogue is considered positive. The negative samples for this dialogue are constructed by replacing some turns with randomly selected turns from other dialogues\footnote{To ensure the generality of our method, we did not use more complex sampling strategies.}. These randomly selected turns bring multiple negative effects (topically in-congruent, semantic confusion, etc.) w.r.t the current dialogue context. However, different from previous works that replaced a fixed number of turns in a $n$>0 rounds dialogue, we set multiple replacement strategies and randomly replace $i\in\{0, 1, ..., n\}$ turns in it. One sample with more replacements is considered of worse overall quality \cite{DBLP:conf/interspeech/GopalakrishnanH19,DBLP:journals/corr/abs-2103-01287}. Specifically, we assign a score $1$ to the original dialogue and assign a score $(n\text{ - }i)/n$ to the new dialogue that replaces $i$ turns. By aligning the replaced turn numbers with a reference score, we get the required training data. Meanwhile, we avoid the quantity and domain limitation of human-annotated data and can easily obtain a large amount of fine-grained training data in different domains. 

Notice that 1) we treat each round with equal weight in this paper, but there may be differences when replacing a turn in the first round (usually a greeting round) or the last one; 2) we hypothesize a linear relationship between the number of replacement turns and the overall dialogue quality, which is not necessarily true. For example, replacing 3 turns and more than 3 turns in the same 6-rounds dialogue may cause the same damage to the overall dialogue quality. We leave these problems for future work.

\begin{figure*}[t]
\centering
\includegraphics[width=\linewidth]{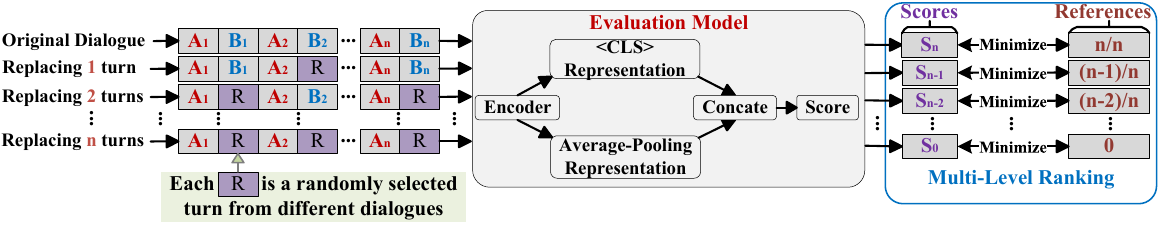}
\caption{The training procedure of SelF-Eval. $\text{A}_i$ and $\text{B}_i$ is the $i$-th turn from speaker A and B, respectively. R is the randomly replaced utterances. The replacement positions can be arbitrary rounds.}
\vspace{-0.5cm}
\label{FigEval}
\end{figure*}

\subsection{Training}
During training, we want to minimize the distance between the predicted score and the reference score. This is a difficult regression task because \textbf{1)} unlike the coherence degree between dialogue context and response or semantic relationships in a Natural Language Inference, our automatic score as regression target lacks clear semantic meaning; \textbf{2)} dialogues with different replaced turns may have the same reference score. For example, the reference score of replacing $1$ turn in a $2$ rounds dialogue is equal to the score of replacing $2$ turns in a $4$ rounds dialogue; \textbf{3)} when the replacement level increases, it is hard for the model to distinguish the small differences. For example, the reference score of replacing $4$ turns in an $8$ rounds dialogue is close to the score of replacing $5$ turns in an $8$ rounds dialogue (0.5 and 0.375, respectively). To smooth the convergence process, we divide the training stage into \textbf{coarse} and \textbf{fine-grained} and name the training stages Multi-level Contrastive learning (MLCL) schema. MLCL is model agnostic and can be used for any similar tasks. 

Figure \ref{FigEval} shows the training process of our model. We choose RoBERTa \cite{DBLP:journals/corr/abs-1907-11692} as encoder. The input for a dialogue sample $\textbf{D}$ is [<CLS>, $A_1$, $B_1$, ..., $A_{n}$, $B_{n}$], where "<CLS>" is a special token and $n$ is different for different $\textbf{D}$. We first obtain two kinds of dialogue representations. The \textit{<CLS>} representation is the output vector $E_{cls}$ of the first token <CLS>. The \textit{Pooling} representation $E_{pooling}$ is obtained by average-pooling all token representations of [$A_1$, $B_1$, $A_2$, $B_2$, ..., $A_{n}$, $B_{n}$]. The final dialogue representation $h_D$ is [$E_{cls}$;$E_{pooling}$], where [;] is the concatenation operation. The $h_D$ is passed through a Multi-layer Perceptron (MLP) to get the predicted quality score $S^{\textbf{D}}$ for $\textbf{D}$:

\vspace{-0.5cm}
\begin{align}
S^{\textbf{D}} \text{ = }  \sigma(W_2\cdot \mu(W_1\cdot h_D \text{ + } b_1) \text{ + } b_2 ),
\end{align}

where $W_{1,2}$ and $b_{1,2}$ are training parameters; $\sigma/\mu$ is the sigmoid/tanh function, respectively.

\subsubsection{The Coarse Training Stage}
Formally, given a training corpus \textbf{C} = ${\{\textbf{D}_m\}}_{m=1}^{M}$ where $\textbf{D}_m$ is the m-th dialogue with $n_m$-rounds. For each $\textbf{D}_m$, we can replace $i\in\{0, 1, ..., n_m\}$ turns and get a replaced version of $\textbf{D}_m$, named $\textbf{D}_m^i$. Each replacement level has its own reference score $(n_m\text{ - }i)/{n_m}$. In the first training stage, we combine a separation loss and a compactness loss as the multi-level ranking (mlr) loss. The mlr loss helps the model learn a coarse granularity ranking ability for multi-levels. 

\textbf{The separation loss} aims to separate the features of different replacement examples by distinguishing their scores. For each replacement level $i\in\{0, 1, ..., n_m\}$, we first calculate a centroid score $S^{\textbf{D}_m^i}$=$\frac{1}{K_i}\sum_{k=1}^{K_i}S_k^{\textbf{D}_m^i}$ where $S_k^{\textbf{D}_m^i}$ is the quality score of a dialogue example with i turns replaced, $K_i$ is the number of contrastive samples for $\textbf{D}_m$ in this replacement level\footnote{For example, when replacing $i$ turns in a $n$ rounds dialogue, we have total $K_i$=$n!/(i!(n\text{ - }i)!)$ contrastive samples.}. The separation loss between different replacement levels is:

\vspace{-0.5cm}
\begin{align}
\mathit{l}_m^{sep} \text{=} \sum_{j=0}^{n_m-1} \sum_{l=j+1}^{n_m} \text{max}(0, \omega \text{*} \lambda \text{+} S^{\textbf{D}_m^j} \text{- } S^{\textbf{D}_m^l}),
\end{align}

where $\lambda$=$1/({n_m}\text{ - }1)$ is the lower bound for the distance between two centroid scores\footnote{For example, $\lambda$=$0.5$ when there are $3$ reference score levels. The expecting 3 centroids are around $1$, $0.5$, and $0$.}, $\omega$ = $l\text{ - }j$ is the weight used for amplifying the lower bound according to the quality-level gap.

\textbf{The compactness loss} aims to compact the examples within the same level, which served as a regularization role to avoid outlier exceptions for each level. Specifically, the dialogue quality score $S_k^{\textbf{D}_m^i}$ for $k\in\{1, 2, ..., K_i\}$ is forced to be closer to the corresponding centroid $S^{\textbf{D}_m^i}$ as follows:

\vspace{-0.5cm}
\begin{align}
\mathit{l}_m^{com} \text{=} &  \sum_{i=0}^{n_m} \sum_{k=1}^{K_i} \text{max}(0,  |S^{\textbf{D}_m^i}\text{ - }S_k^{\textbf{D}_m^i}|\text{ - }\mu),
\end{align}

where $\mu$ is the upper bound for the distance between the centroid of a certain replacing level and the score within this level\footnote{For example, if we set $\lambda$ to $0.3$ and $\mu$ to $0.1$ when there are $3$ reference levels. The expecting 3 level ranges are around [$0.9, 1$], [$0.4, 0.6$], and [$0, 0.1$].}. The mlr loss is:

\vspace{-0.5cm}
\begin{align}
\mathit{L}^{mlr} \text{= } & \sum_{m=1}^{M} (\mathit{l}_m^{seq} \text{ + }\mathit{l}_m^{com}).
\end{align}

The original multi-level ranking method is proposed by \citet{DBLP:conf/acl/YeLHLL20} and has three secondary losses: separation loss, compactness loss, and ordering loss. The difference between the multi-level ranking methods we used and what they used is that we remove the ordering loss and compute the difference instead of the L1 distance between different centroid scores so that the ordering loss is covered by the separation loss. Our method can save training time and keep equal performance.

\begin{table}[t]
\footnotesize

\begin{tabular}{l|c|c|c|c}
\hline

Dataset & \textbf{D}s & turns & words & turns/\textbf{D} \\
\hline
Empathetic & 24,846 & 107,208  & 1.7M & 4.3\\
ConvAI-2       & 18,878 & 278,192  & 3.3M & 14.7\\
DailyDialog          &12,096  &100,360   & 1.4M & 8.3\\
DailyDialog++    & 19,071 & 215,625  & 1.2M & 4.3 \\
GRADE & 1,200 & 2,400 & 61K & 2.0 \\ 
FED    & 500 &  5,603 & 49K & 11.2 \\
DSTC-9 & 2,200 & 59,840 & 533K & 27.2 \\
\hline
\end{tabular}
\caption{Statistics of datasets. "\textbf{D}" means dialogue.}
\vspace{-0.5cm}
\label{statistic}
\end{table}

\subsubsection{The Fine-grained Training Stage}
After the coarse training stage, the model has learned to rank multi-level scores, which can be seen as an approximate fitting to the reference labels. To make the training more smooth and more efficient, we add an R-drop loss \cite{DBLP:conf/nips/LiangWLWMQCZL21} aside from the mlr loss to obtain a more robust representation for each dialogue. The robust representation will help the convergence of the model. Specifically, one input dialogue will go through the model twice and obtain two scores $S_{k, first}^{\textbf{D}_m^j}$ and $S_{k, second}^{\textbf{D}_m^j}$, then the model will minimize the distance between the two scores as follows:

\vspace{-0.5cm}
\begin{align}
\mathit{L}^{drop}\text{ = }& \sum_{m=1}^{M} \sum_{j=0}^{n_m} \sum_{k=1}^{K_i} ( S_{k, first}^{\textbf{D}_m^j}\text{ - }S_{k, second}^{\textbf{D}_m^j})^2,
\end{align}

The overall Loss of the fine-grained training stage $\mathit{L}^{final}$ is computed as follows:

\vspace{-0.5cm}
\begin{align}
\mathit{L}^{final}\text{ = }\mathit{L}^{mlr}\text{ + }\mathit{L}^{drop}.
\end{align}

\section{Experimental Settings}

\subsection{Datasets}
The datasets used in this paper are shown in Table \ref{statistic}. \textbf{Empathetic Dialogue} dataset \cite{DBLP:conf/acl/RashkinSLB19} simulates real life dialogue in which the interlocutor needs to identify and recognize the feelings of others. \textbf{ConvAI-2} \cite{DBLP:conf/acl/KielaWZDUS18,DBLP:journals/corr/abs-1902-00098} mimics the scene where each interlocutor tries to understand each other by incorporating persona information. \textbf{DailyDialog} \cite{DBLP:conf/ijcnlp/LiSSLCN17} reflects our daily communication and covers different topics such as interpersonal relationships and health. \textbf{DailyDialog++} \cite{DBLP:journals/tacl/SaiMAK20} is a multi-reference open-domain dialogue dataset with 3 groups (relevant, irrelevant, and adversarial) of responses for each context, each group has $5$ different responses. \textbf{GRADE} dataset \cite{DBLP:conf/emnlp/HuangYQLL20} contains 300 dialogue examples from Empathetic Dialogue and DailyDialog, and 600 dialogue examples from ConvAI2. Each example has 2 turns with human-annotated relevance scores. \textbf{FED} \cite{DBLP:conf/sigdial/MehriE20} is a set of human-machine and human-human conversations with eighteen fine-grained quality scores in both turn and dialogue levels. \textbf{DSTC-9} \cite{DBLP:journals/corr/abs-2011-06486} was collected on the DialPort platform through direct interaction between real users and open-domain chit-chat systems.

\subsection{Baselines}
We choose the following SOTA models: \textbf{GPT-2} \cite{DBLP:conf/acl/PangNHZLT20} computes the log-likelihood of the response conditional on the the dialogue context normalized by the length of the response; \textbf{QuantiDCE} \cite{DBLP:conf/acl/YeLHLL20} uses BERT \cite{DBLP:conf/naacl/DevlinCLT19} to get dialogue-level representations and proposes a multi-level ranking method to train a quantifiable turn-level coherence metric; \textbf{FED} \cite{DBLP:conf/sigdial/MehriE20} computes the log-likelihood of DialoGPT \cite{DBLP:conf/acl/ZhangSGCBGGLD20} generating predefined positive or negative comments as the quality score. It can measure both turn and dialogue-level qualities; \textbf{DynaEval} \cite{DBLP:conf/acl/ZhangCDZFL020} integrates turn representations from RoBERTa into dialogue-level representation with a graph convolutional network, then adopts contrastive learning to distinguish positive and negative samples. 

We also test with different settings of SelF-Eval. The model shown in Figure \ref{FigEval} is named \textbf{SelF-Eval(full)}, in which the training dialogue can be any rounds. The first different setting is that we use fixed rounds of dialogue data for training. We set all dialogues to 2 rounds and have 3 different replacement strategies: the original dialogue and replacing 1 or 2 turns. This setting is named \textbf{SelF-Eval(simple)}. Besides, we have the following settings for the ablation study. \textbf{SelF-Eval(-mlr)} and \textbf{SelF-Eval(-drop)} means we remove the multi-level ranking loss and D-drop loss, respectively. When removing the mlr loss, we use a binary cross-entropy (BCE) loss instead. It means the model makes a binary decision between original dialogue and dialogue with replacements. We use BCE loss to show our multi-level ranking method is better than a two-level loss when learning a dialogue representation for evaluation.

\subsection{Implementation Details}

The setting of the baseline models follows the papers that proposed them. The pre-trained models (BERT, RoBERTa, DialoGPT, GPT-2) are based on the public Pytorch implementation (https://github.com/huggingface/transformers). The hyper-parameters which are not introduced in this section follow the original implementation in the link. During fine-tuning, we truncate the input dialogue length to $512$ tokens. Among the 7 datasets we used, only DSTC-9 has dialogue examples that exceed 512 tokens and the percentage is 13.6\%. We set the max contrastive sample number to 8. All models are learned with Adam optimizer with $\beta_1$ = $0.9$ and $\beta_2$ = $0.999$. We use a single Tesla A$100$s GPU with $40$GB memory, the batch size is $15$. The average training time for each epoch is around 4 hours (2 hours for the first training stage and 2 hours for the second training stage). The initial learning rate is set to 0.005 and decays to 0.002 in the second stage. A dropout of 0.5 is also applied. When training SelF-Eval(full), the GPU memory occupation is $39$GB. $\mu$ is set to 0.1.\footnote{When evaluating dialogue-level qualities with turn-level metrics, we measure all context-response pairs in a dialogue and use their average as the final score. When evaluating turn-level qualities with dialogue-level metrics, we treat the context-response pair as an entire dialogue.}

\subsection{Evaluation Metrics}
Following previous works \cite{DBLP:conf/sigdial/MehriE20,DBLP:conf/acl/ZhangCDZFL020}, we choose two metrics to correlated with manual evaluations. \textbf{Pearson Correlation} \cite{freedman2007statistics} measures the linear correlation between two sets of data. \textbf{Spearman Correlation} \cite{zar2005spearman} assesses the monotonic relationships between two variables. Besides, we use \textbf{Accuracy} measures the percentage of correct ranking for multi-level replacement.

\section{Experimental Results and Analysis}

We aim to answer the following questions about SelF-Eval: \textbf{(Q1)} \textit{can it assign reasonable scores for multiple replacement levels?} (See section 5.1) \textbf{(Q2)} \textit{does it outperform state-of-the-art methods and truly model the correlations between turns/dialogue?}  (See section 5.2 and 5.3) \textbf{(Q3)} \textit{how do the different components contribute to its performance?} (See section 5.4) \textbf{(Q4)} \textit{what can we learn from case study?} (See section 5.5)

\begin{table}[t]
\footnotesize
\begin{tabular}{l| c|c|c|c}
\hline
model & Rep-0 & Rep-1 & Rep-2 & overall \\
\hline
QuantiDCE            & 0.688 & 0.486& 0.654& 0.609\\
DynaEval                & 0.812 & 0.595&  0.699  & 0.702 \\ 
\hline
SelF-Eval(simple) & {0.962} & {0.891} &  {0.904} & {0.919} \\
SelF-Eval(full) & \textbf{0.973} & \textbf{0.898} &  \textbf{0.914} & \textbf{0.928} \\
\hline
\end{tabular}
\caption{Accuracy of predicting replacement levels.}
\vspace{-0.5cm}
\label{Acc}
\end{table}

\subsection{Ranking Capability (Q1)}

This experiment tests whether an evaluation model assigns higher scores for dialogues with less replacement. Table \ref{Acc} shows the accuracy results of QuantiDCE, DynaEval, and SelF-Eval(simple/full), all models 1) are trained with DailyDialog++ and test with the DailyDialog++ test set; 2) use base-sized pre-trained models as backbones. QuantiDCE is chosen because 1) it is trained for classification and fits perfectly for this experiment; 2) it represents the SOTA turn-level metric. DynaEval is chosen because it is the SOTA dialogue-level metric. We define 3 replacement levels: the original dialogue (Rep-0), the dialogue with 1 replacing turn (Rep-1), and more than 1 replacing turn (Rep-2). Each replacement level has $5010$ samples. 

We can see that SelF-Eval(full) gets the highest performance on all replacement levels. Between the multi-level ranking models, Self-Eval(full) outperforms QuantiDCE by 52.4\%. Between the dialogue-level ranking models, SelF-Eval(full) surpasses DynaEval by 32.2\%. Notably, the accuracy gaps between Rep-(0, 1, and 2) of SelF-Eval(full) are 0.075/-0.016, which are much smaller than the gaps of QuantiDCE (0.202/-0.168) and DynaEval (0.217/-0.104). The results show that 1) SelF-Eval can evenly distinguish the 3 replacement levels; 2) the MLCL method we used shows advantages over multi-level learning in QuantiDCE and contrastive learning in DynaEval.

\begin{table}[t]
\footnotesize
\center
\begin{tabular}{l| c|c|c}
\hline
model & Pearson & Spearman & average \\
\hline
DynaEval                & 0.093 & 0.101  & 0.097 \\ 
FED            & 0.128 & 0.120  & 0.124 \\
\hline
SelF-Eval(simple) & 0.158 &  0.165  &  0.162  \\
SelF-Eval(full) & \textbf{0.163} &  \textbf{0.173}  &  \textbf{0.168}  \\
\hline
\end{tabular}
\caption{Evaluation on DSTC-9 dialogue-level quality.}
\vspace{-0.5cm}
\label{DSTC}
\end{table}

\begin{table*}[t]
\footnotesize
\begin{tabular}{l|c|c|  c|c   |c|c|   c|c|c| c}
\hline
Dialogue Aspects& GPT-2      & Q-DCE & FED        & D-Eval          & S-E(s)    & S-E(f)  & (-drop) & (-mlr) &(-drop,-mlr)& Human \\
\hline
\multicolumn{11}{c}{Dialogue-level (11 quality aspects) Spearman Correlation} \\
\hline
Coherence       & 0.122              & 0.191 & 0.251                & 0.424& 0.423               & \textbf{0.436}& 0.332& 0.340 & 0.137                & 0.809 \\
Error Recovery&\textit{0.097}& 0.109 &  0.165 & 0.351&0.363 & \textbf{0.393}& 0.252& 0.269& 0.135        & 0.840 \\
Consistency     & 0.093              & 0.332 &  \textit{0.116} & 0.326& 0.246              & \textbf{0.347}& 0.233& 0.318& 0.124                &  0.562 \\
Diversity           & 0.145              & \textit{-0.014}&  \textbf{0.420} & 0.342& 0.283            & 0.263& 0.197& 0.116&  \textit{0.022} &  0.789 \\
Topic Depth     &\textit{0.094} & \textit{-0.054}&  \textbf{0.476} & 0.375& 0.316            & 0.327& 0.204& 0.177&  \textit{0.004} & 0.833 \\
Likability          & 0.178              & 0.098  & 0.262                & 0.357& 0.345   & \textbf{0.390}& 0.285& 0.275&\textit{0.074}& 0.838 \\
Understanding  &\textit{0.073} & 0.210  & 0.306               & 0.373& 0.364                &\textbf{0.406}& 0.329& 0.306&  0.108 & 0.809 \\
Flexibility           & 0.135             & 0.093  & 0.293               &\textbf{0.361} & 0.307& 0.317& 0.233& 0.184&  \textit{0.082}                 &  0.816 \\
Informativeness& 0.119            & \textit{0.063}  & 0.288                & 0.302& 0.311 & \textbf{0.318}& 0.184& 0.194&  \textit{0.019} &  0.806 \\
Inquisitiveness  &\textit{0.070} & 0.115& 0.163               & 0.294& 0.401 & \textbf{0.421}& 0.309& 0.267&  \textit{0.085} &  0.769 \\
\hline
Overall                &0.121 & 0.140  &\textbf{0.443} & 0.428& 0.428                & 0.435& 0.252& 0.272&  \textit{0.055} &  0.830 \\

\hline

\multicolumn{11}{c}{Turn-level (9 quality aspects) Spearman Correlation} \\
\hline
Interestingness         & -0.097         & -0.163            & \textbf{0.408} & 0.197         &0.146 &0.183                         & 0.143 & 0.142 & 0.063 & 0.819 \\
Engagement             & \textit{-0.096}& -0.138       & \textbf{0.318} & 0.119        &0.149   &0.206                         & 0.126 &0.128&0.047& 0.798 \\
Specificity                 & -0.114          & -0.171            & \textbf{0.267} & 0.161         &0.097 &0.169                         & 0.139  &0.112 & \textit{0.069} & 0.790 \\
Relevance                 & -0.103           & \textit{0.085} & 0.152             & 0.171         &0.263 &\textbf{0.282}& 0.174&0.175&0.101& 0.753 \\
Correctness             & \textit{0.041} & \textit{0.080} & 0.133           & 0.165          &0.233 &\textbf{0.291}& 0.165&0.163&\textit{0.083}& 0.780 \\
\textbf{S}.Appropriateness&\textit{-0.081} &0.112 & 0.155          & 0.112          &0.208 &\textbf{0.247}& 0.143&0.132&\textit{0.068}& 0.682 \\
Understandable     & \textit{-0.076} & 0.195             & 0.111           & 0.116            &0.136 &\textbf{0.173}& 0.110&0.111&\textit{0.046}& 0.522 \\
Fluency                    & -0.154             &\textit{0.071}&\textbf{0.224}&\textit{0.016} &0.095 &\textit{0.038}       & \textit{0.034}&\textit{0.036}&\textit{-0.047}& 0.714 \\
\hline
Overall                    & \textit{-0.090}  & \textit{0.014}                 & 0.209                 & 0.207            &0.255 &\textbf{0.292}&0.146&0.140&\textit{0.067}& 0.820 \\

\hline
\end{tabular}
\caption{Comparison of both dialogue and turn level Spearman correlations on the FED evaluation dataset. Q-DCE/D-Eval/S-E(s)/S-E(f)/"\textbf{S}." is short for QuantiDCE/DynaEval/SelF-Eval(simple)/SelF-Eval(full)/Semantically, respectively. Scores with p-values larger than 0.01 are italicized (indicating statistical insignificance).} 
\vspace{-0.5cm}
\label{FED}
\end{table*}

\subsection{Experiments on DSTC-9 (Q2)}
Table \ref{DSTC} shows the experimental results on DSTC-9 data. The DSTC-9 dataset is difficult to evaluate because of two reasons: 1) it contains direct interaction between real users and multiple open-domain chit-chat systems. These chit-chat systems are trained with dialogue data in different domains compared with ours. In another word, the DSTC-9 dialogue data is out-of-domain for our model and can be used to test the generality of our method; 2) the average turns in a dialogue is around 27.2, which is the longest among all datasets we used and also much longer than the training data we used. It is difficult for evaluation models to give a score for such a long conversation. In this experiment, all models are fine-tuned on DailyDialog++ and the RoBERTa-based models are all first pre-trained on Empathetic Dialogue, ConvAI-2, and DailyDialog. 

Pearson and Spearman correlations between the model-generated scores and the corresponding human evaluation scores are computed in Table \ref{DSTC}. FED and DynaEval are chosen because they are the SOTA dialogue-level evaluation models. We can see that both SelF-Eval(simple) and SelF-Eval(full) largely outperform SOTA baselines even though all models are affected by the out-of-domain and long conversation problems. The results show that SelF-Eval is capable of learning a dialogue representation for evaluating dialogue-level quality even in a difficult dataset such as DSTC-9. To further verify the generality of this evaluation ability, we test with other out-of-domain datasets in the following sections.

\subsection{Experiments on FED (Q2)}
Table \ref{FED} shows the experimental results on FED data. In both dialogue and turn-level evaluations, Spearman correlations between the model-generated scores and the corresponding human evaluation scores are computed. Models are trained in the same setting as experiments on DSTC-9. 

\subsubsection{Dialogue-level Evaluation}
There are 11 different aspects of the FED dialogue-level evaluation. GPT-2 and QuantiDCE are SOTA turn-level evaluation metrics. They evaluate a dialogue based on the aggregation of scores of all the context-response pairs within the dialogue. We can observe that most of their correlation scores (21 out of 22) on dialogue aspects are lower than those of FED and DynaEval. The results are consistent with the conclusion of previous studies \cite{DBLP:journals/corr/abs-2106-03706} that turn-level quality evaluation may be insufficient to assess the dialogue-level performance. 

FED has the highest scores on Diversity, Topic Depth, and Overall. These results may indicate that the DialoGPT-based evaluation model (FED) is better at measuring these three attributes than the RoBERTa-based models (DynaEval and SelF-Eval). The reason is that DialoGPT uses a large amount of Reddit data for training. The diverse topics and variation expressions in Reddit data provide DialoGPT with more insights on these attributes, especially the dialogue-level Overall attribute. In contrast, DynaEval and SelF-Eval are trained with fewer dialogue data (fewer topics and variation expressions). The DynaEval focuses on modeling the dependency between pairs of utterances and the SelF-Eval focuses on modeling the correlations between turns and the entire dialogue. They are more useful for evaluating Coherence, Error Recovery, and Consistency aspects which reflect the interaction between turns. Specifically, SelF-Eval owns the highest correlation scores in 7 out of 11 dialogue aspects (Coherence, Error Recovery, Consistency, likability, Understanding, informativeness, and Inquisitiveness) and the second-highest correlation scores on Flexibility and Overall. SelF-Eval successfully learns to measure these attributes with our replacement strategies. The MLCL training method captures the various dialogue attributes and entails good dialogue-level representations. The dialogue-level evaluation tasks are benefiting from this representation. One way to improve the Diversity and Topic Depth scores of RoBERTa-based models is to pre-train them with dialogue data that contains more topics and domains. We can also notice that SelF-Eval(full) is better than SelF-Eval(simple) in most aspects. It means the simplified training method used by SelF-Eval(simple) is not as strong as the original method introduced in Figure \ref{FigEval}.

\subsubsection{Turn-level Evaluation}
There are 9 different aspects of the FED turn-level evaluation. The turn-level metrics (GPT-2 and QuantiDCE) only get better correlations on 6 out of 18 aspects than the dialogue-level metrics (FED and DynaEval). The results indicate that the generality of these two turn-level evaluation models is not strong. They work well only in constrained environments or on specific datasets. The FED model achieves the highest correlation on Interestingness, Engagement, Specificity, and Fluency. The reason is that the DialoGPT used by FED is trained with an auto-regressive mode and models language generation word by word. DialoGPT focuses more on the token-level correlations and is effective for evaluating the naturalness of an utterance. In contrast, all the RoBERTa-based models (DynaEval and SelF-Eval) perform poorly for token-level aspects. This is because they focus on the correlations in the turn level and do not pay enough attention to the token level. One way to strengthen the fluency and Specificity aspects of SelF-Eval is to introduce token-level perturbation strategies in training data, such as word drop and addition \cite{DBLP:conf/emnlp/SaiDSMK21}. These strategies provide negative samples with semantical or grammatical mistakes which may also be used for setting multi-level turn qualities for training. We consider this token-level perturbation as future work. What's more, we have a similar finding to \citet{DBLP:conf/acl/ZhangCDZFL020} that SelF-Eval(s) and FED complement each other at turn-level. It means that they both perform well in aspects that the other one is not good at. SelF-Eval achieves the highest correlation in Relevance, Correctness, Semantically Appropriateness, Understandable, and Overall. The SelF-Eval(simple)/(full) outperforms the best baseline  39.7\%/22.0\% on turn-level Overall, respectively. It also has the second-highest scores on Engagement and Specificity. The results are consistent with the dialogue-level evaluation where SelF-Eval has good results on aspects that reflect the interaction between turns. As in the dialogue-level, SelF-Eval(full) is still better than SelF-Eval(simple) on most aspects in turn-level. 

To sum up the experiments on the FED dataset, SelF-Eval performs well for both dialogue-level and turn-level evaluations, especially the latter. The reason is that the training process of SelF-Eval not only models the correlation between turns and the entire dialogue but also models the inner connection between context and response. Our method successfully aligns the semantic information shared by turns and dialogue and shows good domain adaptability (on both DSTC-9 and FED). 

\subsection{Ablation Study (Q3)}

Table \ref{FED} also shows the ablation study of the SelF-Eval(full). Removing R-drop loss (-drop) in the fine-grained training stage causes more declines than (-mlr) in dialogue-level evaluations. This is because the R-drop loss helps SelF-Eval to learn more robust dialogue representations. Replacing the mlr loss with BCE loss (-mlr) causes more declines than (-drop) in turn-level evaluations. This indicates that the mlr loss helps to distinguish the turn replacement levels and the semantic inconsistency information caused by the replacements. When removing both mlr and R-drop losses, the performance declines significantly and the results become statistical insignificance.

\subsection{Case Study (Q4)}
\begin{table}[t]
\footnotesize
\begin{tabular}{l}
\hline
\textbf{U1}: My partner left me the other day. \\
\textbf{U2}: That's rough, I'm sorry to hear that. \\
\textbf{R}: Being a punching bag in a relationship is no good. \\
\ \ \ \ It's a 2 way street. Is your partner doing their part?\\
\textbf{Scores}(Human / \textcolor{red}{SelF-Eval} / \textcolor{blue}{DynaEval}): 0.77 / \textcolor{red}{0.84} / \textcolor{blue}{0.50}\\
\hline
\textbf{U1}: i was so stressed when i found out that i did not get \\
\ \ \ \ accepted in my dream college.\\
\textbf{U2}: Oh no. Did you have a good backup plan?\\
\textbf{R}: thats cool, i hope you have a good time.\\
\textbf{Scores}(Human / \textcolor{red}{SelF-Eval} / \textcolor{blue}{DynaEval}): 0.25 / \textcolor{red}{0.09} / \textcolor{blue}{0.46}\\

\hline
\end{tabular}
\caption{Case study on GRADE. U1/U2 are the dialogue context and R is the response to be evaluated.}
\vspace{-0.5cm}
\label{CS}
\end{table}

We randomly select 2 examples from GRADE for the case study (Table \ref{CS}). The task is to evaluate the response when giving dialogue context. We compare the human rating (the relevance scores) with the scores given by DynaEval and SelF-Eval(full). They are both based on RoBERTa and could provide more insight into our model. In both cases, the scores given by SelF-Eval are closer to the human rating score than DynaEval. This is consistent with the experimental results in the turn-level Relevance of the FED dataset. In both cases, scores from SelF-Eval are more polarized than the human evaluations. This indicates that humans may be reluctant to give extreme scores and SelF-Eval could improve its performance by penalizing scores that are too extreme. However, whether this penalizing works for dialogue attributes other than relevance requires further study.

\section{Conclusion}
We propose to measure dialogue quality by modeling the fine-grained correlations between turns and the entire dialogue. We introduce our data construction method and SelF-Eval model. Experiments show that SelF-Eval builds fine-grained correlations and gets better correlation scores with human ratings than SOTA models. We think our method may have two potential applications: 1) It can be used alone in the evaluation of dialogue tasks after training the SelF-Eval model with a large amount of in-domain data; 2) It can be combined with other evaluation models (such as FED) to evaluate the dialogue task by integrating the advantages of different evaluation models. In the future, we would like to improve our method by 1) employing multi-granularity turn-level scores; 2) modeling the nonlinear relationships between replacement numbers and dialogue quality. 

\section{Acknowledgments}
This paper is supported by the Science and Technology Innovation 2030 Major Project of China (No. 2021ZD0113302) and National Natural Science Foundation of China (No. 62076081, No. 61772153 and No. 61936010).

\bibliography{acl_latex}

\end{document}